\documentclass{article} 
\usepackage[preprint]{colm2025_conference}

\usepackage{microtype}
\usepackage{hyperref}
\usepackage{url}
\usepackage{booktabs}
\usepackage{graphicx}
\usepackage{wrapfig}
\usepackage{amsmath}
\usepackage{multirow}

\usepackage{lineno}

\definecolor{darkblue}{rgb}{0, 0, 0.5}
\hypersetup{colorlinks=true, citecolor=darkblue, linkcolor=darkblue, urlcolor=darkblue}

\title{Improved Visual-Spatial Reasoning via R1-Zero-Like Training}

\author{\small Zhenyi Liao$^1$\thanks{Works done during the internship in OPPO AI Center.},\;Qingsong Xie$^2$\thanks{Corresponding authors.}, Yanhao Zhang$^2$, \;Zijian Kong$^1$,Haonan Lu$^2$, Zhenyu Yang$^2$,\;Zhijie Deng$^1$\footnotemark[2] \\
$^1$Shanghai Jiao Tong University \quad $^2$OPPO AI Center \\
\texttt{\{l-justice, kong-zijian, zhijied\}@sjtu.edu.cn
}\\
\texttt{\{xieqingsong,zhangyanhao,luhaonan,yangzhenyu\}@oppo.com}
}

\begin{document}
 
\ifcolmsubmission
\linenumbers
\fi

\maketitle

\begin{abstract}
Increasing attention has been placed on improving the reasoning capacities of multi-modal large language models (MLLMs). 
As the cornerstone for AI agents that function in the physical realm, video-based visual-spatial intelligence (VSI) emerges as one of the most pivotal reasoning capabilities of MLLMs.
This work conducts a first, in-depth study on improving the visual-spatial reasoning of MLLMs via R1-Zero-like training. 
Technically, we first 
identify that \emph{the visual-spatial reasoning capacities of small- to medium-sized Qwen2-VL models cannot be activated via Chain of Thought (CoT) prompts}. 
We then incorporate GRPO training for improved visual-spatial reasoning, using the carefully curated \textbf{VSI-100k} dataset, following DeepSeek-R1-Zero. 
During the investigation, we 
identify the necessity to \emph{keep the KL penalty (even with a small value) in GRPO}. 
With just 120 GPU hours, our \textbf{vsGRPO-2B} model, fine-tuned from Qwen2-VL-2B, can outperform the base model by 12.1\% and surpass GPT-4o. 
Moreover, our \textbf{vsGRPO-7B} model, fine-tuned from Qwen2-VL-7B, achieves performance comparable to that of the best open-source model LLaVA-NeXT-Video-72B.
Additionally, we compare vsGRPO to supervised fine-tuning and direct preference optimization baselines and observe strong performance superiority.
The code and dataset will be available at \url{https://github.com/zhijie-group/R1-Zero-VSI}.

\end{abstract}

\section{Introduction}
Multimodal Large Language Models (MLLMs) have emerged as a significant advancement in AI. 
They typically accept text, images, and videos as inputs and output textual responses, serving as the foundations for various applications, including multi-modal understanding~\citep{liu2023visual,wang2024qwen2}, visual language agents~\citep{hong2024cogagent,wang2024videoagent}, autonomous driving~\citep{pan2024vlp,wang2024driving}, etc. 

The exhaustive understanding of multi-modal observations hinges on advanced reasoning capabilities, which has spurred growing interest in investigating reasoning mechanisms within MLLMs~\citep{suris2023vipergpt,xu2024llava,zhang2025r1}. 
This trend mirrors concurrent advancements in vanilla LLMs~\citep{yao2023tree,lightman2023let,wei2022chain,wangself}. 
As the foundation for AI agents operating in the physical world, the video-based visual-spatial reasoning stands out as one of the most crucial capacities of MLLMs. 
Yet, existing models often fall short in this (see VSI-bench~\citep{yang2024thinking}).

This work conducts a systematic study on improving the visual-spatial reasoning capacities of MLLMs based on R1-Zero-like training. 
Focusing on the Qwen2-VL models~\citep{wang2024qwen2}, we first perform an initial study on whether simple reasoning-oriented prompts can activate the visual-spatial reasoning capacities. 
We investigate various CoT~\citep{wei2022chain} strategies, but find that \textbf{vanilla non-CoT prompts perform the best for small- to medium-sized Qwen2-VL on VSI-bench}. 
This exposes the issue that such models does not have the ability to trade inference FLOPs for improved visual-spatial reasoning. 

Following the journey of DeepSeek-R1-Zero~\citep{guo2025deepseek}, we decide to improve the visual-spatial reasoning capacities of Qwen2-VL with GRPO~\citep{shao2024deepseekmath}, sharing a similar task with the very recent work VisualThinker-R1-Zero~\citep{zhou2025r1}.
Considering the scarcity of training data for visual-spatial question answering, we construct a video-based question answering dataset of more than 100k samples, \textbf{VSI-100k}, following the protocol of VSI-bench. 
Specifically, we leverage ScanNet~\citep{dai2017scannet} to get high-fidelity video scans accompanied by meticulous object-level 3D annotations, based on which (question, answer) pairs regarding spatial information can be easily crafted.

Following common practice~\citep{guo2025deepseek,chen2025r1v,zhou2025r1}, we define the rule-based reward function based on the alignment between the model prediction and the ground-truth answer to perform GRPO. 
We also include a format reward when trying to activate the CoT reasoning behaviour. 
The GRPO on VSI-100k turns the pretrained Qwen2-VL-2B model into the performant \textbf{vsGRPO-2B} within just 120 GPU hours.
We observe that vsGRPO-2B outperforms the base model by 12.1\%.
The same pipeline also transforms the Qwen2-VL-7B model into \textbf{vsGRPO-7B}, achieving performance similar to that of the best open-source model with 72B parameters.
During GRPO training, we have identified the necessity to \textbf{keep the KL penalty (even with a small value) in the training of GRPO} and observed phenomena such as reward hacking. 
We also compare GRPO with supervised fine-tuning (SFT) and direct preference optimization (DPO)~\citep{rafailov2023direct}, and confirm the superiority of GRPO in improving the spatial-visual reasoning capacities of Qwen2-VL.

\section{Can Visual-spatial Reasoning Capacities Be Activated by Prompting?}

We initiate by evaluating Qwen2-VL on the VSI-bench with various prompting strategies.  

Concretely, the VSI-bench includes two types of question-answer problems:
\begin{itemize}
    \item Numerical Answer (NA), including tasks such as object counting, absolute distance measurement, object size evaluation, and room size assessment;
    \item Multiple-Choice Answer (MCA), including tasks related to relative distance, relative direction, route planning, and appearance order.
\end{itemize}

\begin{table}[t]
    \centering
    \setlength{\tabcolsep}{1.3mm}
    \renewcommand{\arraystretch}{1.1} 
    \begin{tabular}{lcccccccccc}
    \toprule
     Backbone & Methods & Avg & \shortstack{ Obj.\\ Count} &  \shortstack{Abs. \\Dist.} & \shortstack{Obj.\\Size} & \shortstack{Room \\Size}
    & \shortstack{Rel. \\Dist.}& \shortstack{ Rel.\\ Dir.}&\shortstack{Route\\ Plan} & \shortstack{Appr.\\ Order}  \\
    \midrule
    \multirow{3}{*}{\parbox{1.5cm}{Qwen2-\\VL-2B}} 
    &{\color{red} Think-mode} & 22.9 & 18.4 & 4.3 & 31.5 & 17.3 & 28.3 & 22.9 & 26.2 & 16.8 \\ 
    &{\color{green} Observe-mode}  & 21.8 & 16.8 & 1.7 &32.7 & 22.7 & 28.8 & 27.6 & 26.2 & 18.1 \\ 
    &{\color{blue} Vanilla-mode}  & 23.3 & 21.4 & 3.4 &32.3 & 31.1 & 26.7 & 27.7 &24.7 &18.9 \\
    \midrule
    \multirow{3}{*}{\parbox{1.5cm}{Qwen2-\\VL-7B}} 
    &{\color{red} Think-mode} & 31.3 & 44.8 & 26.1 &25.3 & 23.4 & 34.7 & 30.9 &32.9 & 31.5  \\ 
    &{\color{green} Observe-mode}  & 32.0 & 29.9 & 19.0 &39.6& 32.0 & 34.6 & 40.0 & 36.0 &24.4   \\ 
    &{\color{blue} Vanilla-mode}  & 32.2 & 39.4 & 25.0 & 25.8 & 43.2 & 32.6 & 30.9 &27.8 & 32.6  \\
    \bottomrule
    \end{tabular}
    \caption{Quantitative comparisons of different prompting strategies on Qwen2-VL-2B and Qwen2-VL-7B on VSI-bench. }
    \label{tab:prompting}
\end{table}

To evaluate the reasoning capacities of Qwen2-VL on this dataset, we consider two CoT prompting strategies: the widely adopted {\color{red} think-mode}, where the model first thinks and then replies to the question, and the {\color{green} observe-mode}, where the model first observes the input video and then replies. 
The latter follows a human-like pattern and has been explored in related works~\citep{wu2024dettoolchain}.
We also include a non-CoT {\color{blue} vanilla-mode} for comparison. 
Here is a summarization of them: 
\begin{itemize}
    \item {\color{red} Think-mode}: \texttt{Let's think step by step and then answer the question using a single word or phrase.}
    \item {\color{green} Observe-mode}: \texttt{Please observe the video first and then answer the question using a single word or phrase.} 
    \item {\color{blue} Vanilla-mode}: \texttt{Please answer the question using a single word or phrase.}
\end{itemize}

\textbf{CoT prompting is ineffective for small- to medium-sized Qwen2-VL on VSI-bench.}
As shown in Table~\ref{tab:prompting}, despite longer responses, {\color{red} think-mode} and {\color{green} observe-mode} underperform the simple {\color{blue} vanilla-mode}.
Namely, small- to medium-sized Qwen2-VL cannot trade inference FLOPs for improved visual-spatial reasoning.

\begin{figure}[t]
    \centering
\includegraphics[width=\linewidth]{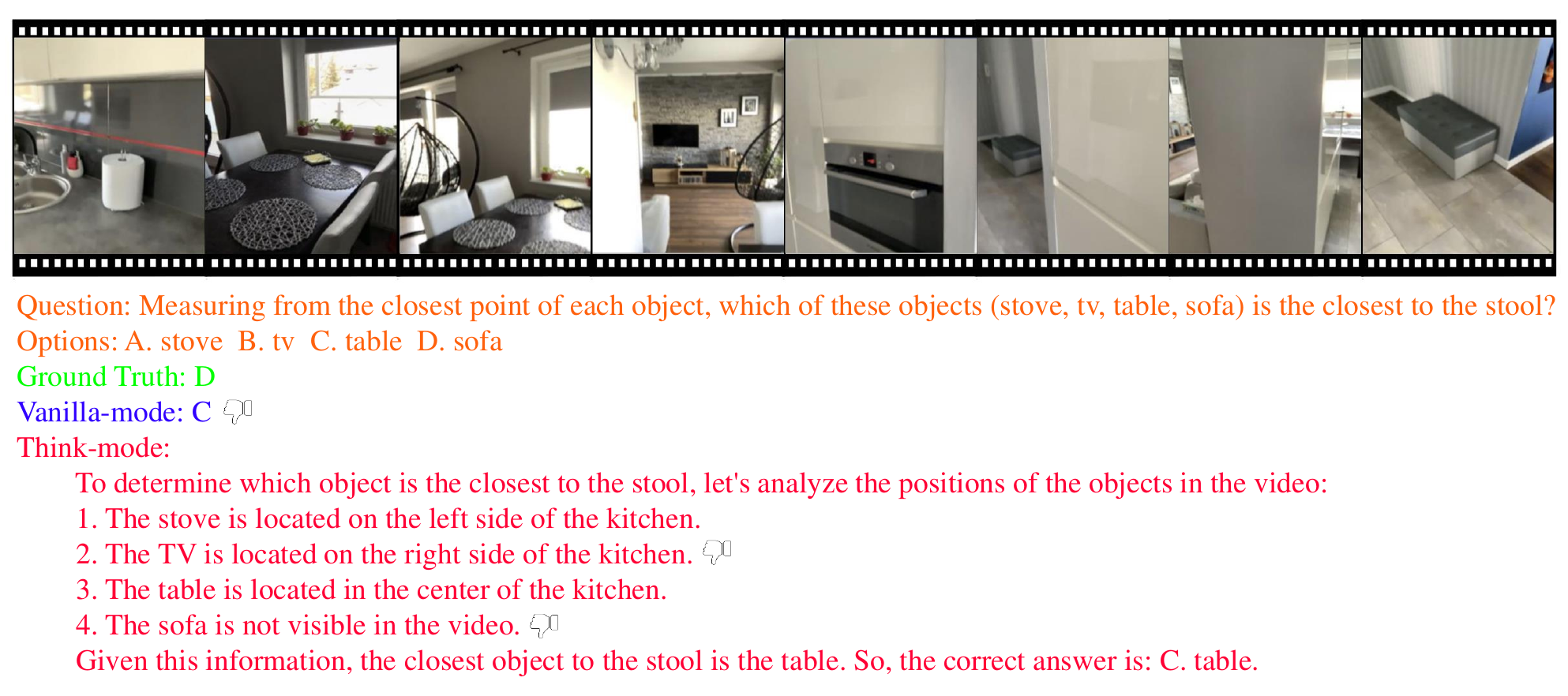}
    \caption{Comparison between the {\color{blue} vanilla-mode} and {\color{red} think-mode} predictions.}
    \label{fig:info_loss}
\end{figure}

We visualize some output examples given by Qwen2-VL-2B in Figure~\ref{fig:info_loss}. 
We see that the model can actually understand the instructions for activating thinking, but the final answer is still wrong, the same as that of the vanilla prompting.
From the exposed chain of thoughts, we realize that the error may arise from the failure to perceive the sofa in the video.

\section{R1-Zero-like Training for Visual-spatial Reasoning}
Given the above observations, we realize it is necessary to fine-tune the Qwen2-VL models for improved visual-spatial reasoning. 
Typically, we opt to focus on the GRPO approach~\citep{shao2024deepseekmath} given its success in building DeepSeek-R1-Zero. 

\subsection{Training Data Construction}

\begin{figure*}[t]
    \centering
    \includegraphics[width=\linewidth]{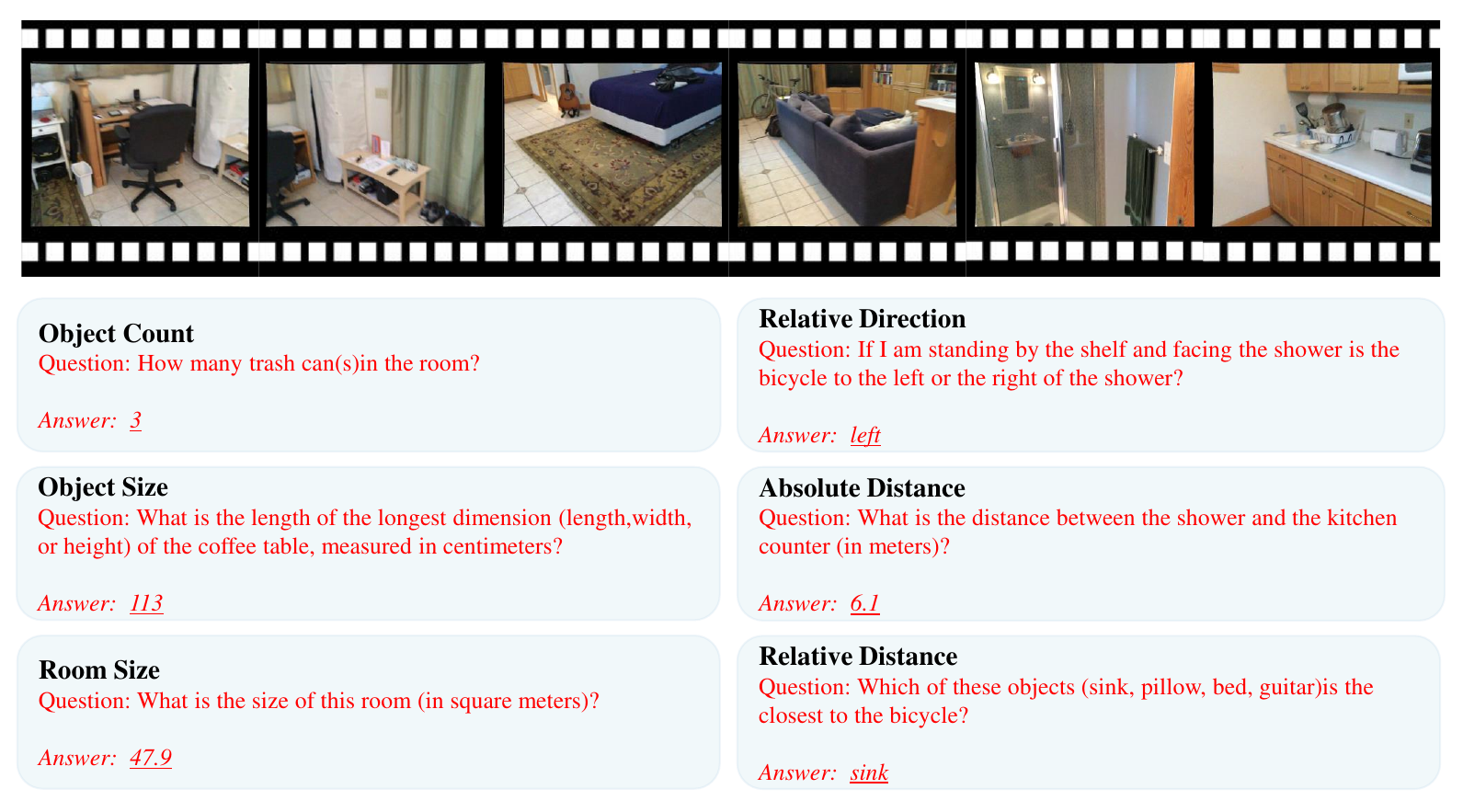}
    \caption{Illustrations of the constructed dataset.}
    \label{fig:dataset}
\end{figure*}

We first create a video-based question-answering dataset named \textbf{VSI-100k} for visual-spatial reasoning. 
It consists of more than 100k samples and follows the VSI-bench protocol.
Specifically, we utilize ScanNet~\citep{dai2017scannet} to obtain high-fidelity video scans that come with detailed object-level 3D annotations. With such information, it becomes straightforward to generate (question, answer) pairs related to spatial information. 

Specifically, we construct questions regarding six topics, including object count, relative direction, relative distance, object size, room size, and absolute distance.
We leave the other two topics in VSI-bench, route planning and appearance order, held out. 
This corresponds to two reasons: 1) (question, answer) pairs of the latter two topics cannot be simply constructed given rarely the static 3D information, which implies that expensive manual annotation can be required; 2) with this, we can test the task generalization ability of the trained models. 
For the NA type problems, we implement a question template similar to that used in~\citep{yang2024thinking}. 
For the MCA one, we simplify the question format by removing the options.
This adjustment enhances the model's capacity to recognize entity correspondence instead of simply matching symbols.
Some examples are provided in Figure~\ref{fig:dataset}. 

For object count, we construct answers with directly the object labels included in the annotations document, yielding a total of 6.4k samples. 
For absolute distance, we first remove objects that appear multiple times to ensure specification preciseness, and then calculate the distance between geometric centers of various 3D point-cloud objects, obtaining 75k samples. 
For relative distance, we fix one targeted object and compute the absolute distance between it and four other objects to estimate relative distance, yielding 13k samples. 
For relative direction, we select one object as the front and determine the relative direction of two objects based on their geometric centers of point clouds, getting a total of 8k samples. 
For object size, we leverage the 3D bounding box to compute the longest dimension of the object, yielding a total of 13k samples. 
For room size, we apply the alpha shape algorithm to the total 1.5k scenes, resulting in 1.5k samples. 

\subsection{GRPO}
Group Relative Policy Optimization, also abbreviated as GRPO~\citep{shao2024deepseekmath}, is a type of reinforcement learning (RL) that eliminates the critic model to reduce training costs.
Specifically, a group of generated output set $\{o_1,o_2,\cdots,o_G\}$ is sampled for each question $q$ from policy model $\pi_{\theta_{old}}$.
Then GRPO optimizes the model $\pi_{\theta}$ using the following objective:  
\begin{equation}
\small
\begin{aligned}
J_{\mathrm{GRPO}}(\theta) 
&= \mathbb{E}_{q \sim P(Q),\, \{o_i\}_{i=1}^G \sim \pi_{\theta_{\mathrm{old}}}(O \mid q)} \\
&\Biggl[
  \frac{1}{G} \sum_{i=1}^G
  \min\!\Bigl(
    \frac{\pi_{\theta}(o_i \mid q)}{\pi_{\theta_{\mathrm{old}}}(o_i \mid q)}\,A_i,\, 
    \mathrm{clip}\!\Bigl(
      \frac{\pi_{\theta}(o_i \mid q)}{\pi_{\theta_{\mathrm{old}}}(o_i \mid q)},
      1-\varepsilon,\,
      1+\varepsilon
    \Bigr)
    A_i
  \Bigr)  
  -\,\beta\,D_{\mathrm{KL}}\bigl(\pi_{\theta}\,\big\|\,\pi_{\mathrm{ref}}\bigr)
\Biggr],
\end{aligned}
\label{eq1}
\end{equation}
where $\varepsilon$ and $\beta$ are the clipping hyper-parameter and the coefficient controlling the Kullback–Leibler (KL) penalty, respectively, and $A_i = \frac{r_i - \operatorname{mean}(\{r_1, r_2, \dots, r_G\})}{\operatorname{std}(\{r_1, r_2, \dots, r_G\})}$ is the computed advantage using the group rewards $\{r_1,r_2,\cdots,r_G\}$.
$D_{KL}\bigl(\pi_\theta \,\|\, \pi_{\text{ref}}\bigr)
= \frac{\pi_{\text{ref}}(o_i \mid q)}{\pi_\theta(o_i \mid q)}
- \log\!\Bigl(\frac{\pi_{\text{ref}}(o_i \mid q)}{\pi_\theta(o_i \mid q)}\Bigr)
- 1 \,$ is the KL divergence.

The reward $r$ guides the direction of the training process and is crucial. 
We adhere to~\citep{chen2025r1v,meng2025mm} of using format rewards and accuracy rewards, but with necessary modifications. 

\noindent\textbf{Format Reward.}
Although the CoT prompts are useless for the small-sized Qwen2-VL-2B in inference time, we still wonder if training with them is beneficial for GRPO. 
As a result, following recent progress in the community, we consider three training prompts for GRPO:
\begin{itemize}
    \item {\color{red}Think-mode}: \texttt{Please think step by step and enclose your thinking process in <think> </think> tags and then provide the short answer with one or two words or a number in <answer> </answer>.}
    \item {\color{green}Observe-mode}: \texttt{Please observe carefully and analyze what you see that helps you to solve the question in the video and enclose it in <observe> </observe> tag, and then provide the short answer with one or two words or a number in <answer> </answer>.}
    \item {\color{blue}Vanilla-mode}: \texttt{Please provide the short answer with one or two words or a number.} 
\end{itemize}

The format reward quantifies how the responses follow the specified format.
It returns a score of 1 or 0.
Note that such a reward is omitted for the {\color{blue}vanilla-mode}.

\noindent\textbf{Accuracy Reward.}
In the case of non-NA tasks, we employ a character matching method to assess accuracy, awarding a score of 1 for a match and 0 for a mismatch. 
For NA tasks, we develop a function that computes the absolute difference between the true value and the predicted one and divides the result by the minimum of the two values.

\begin{table}[t]
    \centering
    \setlength{\tabcolsep}{1.5mm}
    \renewcommand{\arraystretch}{1.1} 
    \begin{tabular}{lcccccccccc}
    \toprule
     Methods & \shortstack{Eval.\\ Mode} &Avg & \shortstack{ Obj.\\ Count} &  \shortstack{Abs. \\Dist.} & \shortstack{Obj.\\Size} & \shortstack{Room \\Size}
    & \shortstack{Rel. \\Dist.}& \shortstack{ Rel.\\ Dir.}&\shortstack{Route\\ Plan} & \shortstack{Appr.\\ Order}  \\
    \midrule
    \multicolumn{11}{c}{\textbf{Open-source}} \\
    \midrule
    Qwen2-VL-2B  & \textcolor{blue}{V} & 23.3 & 21.4 & 3.4 &32.3 & 31.1 & 26.7 & 27.7 &24.7 &18.9 \\
     + SFT & \textcolor{blue}{V} & 29.6 & 29.6 & 23.5 &47.4 & 33.5 & 26.9 & 28.3 &28.8 & 18.6  \\
    + DPO  & \textcolor{blue}{V}& 23.9 & 21.7 & 3.7 & 34.8 & 32.4 & 27.1 & 28.5 & 24.2 & 18.6 \\
    + vsGRPO-T & \textcolor{blue}{V}
    & 26.1 & 24.7 & 10.7 & 37.4 & 36.2 & 27.3 & 29.5 &25.7& 17.9 \\
    + vsGRPO-O & \textcolor{blue}{V}& 28.0 & 26.2 & 16.4 &44.8 & 38.2 & 27.0 & 29.3 &24.2 & 18.2 \\
    + vsGRPO-T& \textcolor{red}{T} & 29.6 & 35.0 & 28.2 & 34.7 & 25.2 & 28.0 & 38.5 & 28.5 & 18.7 \\
   + vsGRPO-O& \textcolor{green}{O}  & 31.2 & 34.6 & 22.5 &44.8 & 33.7 & 29.4 & 41.8 &26.8 & 15.8 \\
   + vsGRPO-V &\textcolor{blue}{V}& \underline{35.4} & 53.6 & 29.0 & 52.7 & 43.4 & 28.1 & 30.9 & 26.8 & 18.9 \\
   \midrule
    Qwen2-VL-7B  & \textcolor{blue}{V} & 32.2 & 39.4 & 25.0 & 25.8 & 43.2 & 32.6 & 30.9 &27.8 & 32.6 \\
     + SFT & \textcolor{blue}{V} & 38.1 & 44.7 & 27.6&46.1&50.4 & 34.0 & 35.7 &33.0& 33.4  \\
    + DPO  & \textcolor{blue}{V}& 32.6 & 39.1 & 25.2 & 26.5 & 44.2 & 32.6 & 30.9 & 29.3 & 33.3 \\
   + vsGRPO-V &\textcolor{blue}{V}& \underline{40.7} & \textbf{59.9} & \textbf{29.6} & 50.8 & 48.3 & 35.4 & 35.6 & 34.0& 31.5 \\

   \midrule
   IVL2-2B &\textcolor{blue}{V} & 27.4 & 21.8 & 24.9 & 22.0 & 35.0 & 33.8 & 44.2 & 30.5 & 7.1\\
   LNV-7B & \textcolor{blue}{V} & 35.6 & 48.5 & 14.0 & 47.8 & 24.2 & 43.5 & 42.4 & 34.0 & 30.6\\
   IVL2-40B  &\textcolor{blue}{V} &36.0& 34.9 & 26.9 & 46.5 & 31.8 &42.1 &32.2 & 34.0 & 39.6   \\
    LNV-72B &\textcolor{blue}{V} & \textbf{40.9} & 48.9 & 22.8 & 57.4 & 35.3 & 42.4 & 36.7 & 35.0 & 48.6    \\
    \midrule
    \multicolumn{11}{c}{\textbf{Close-source}} \\
    \midrule
     GPT-4o  &\textcolor{blue}{V} &34.0& 46.2 & 5.3 & 43.8 & 38.2 &37.0 & 41.3 & 31.5 & 28.5   \\
    Gemini-1.5 Pro &\textcolor{blue}{V} & \textbf{48.8} & 49.6 & 28.8 & \textbf{58.6} & \textbf{49.4} & \textbf{46.0} & \textbf{48.1} & \textbf{42.0} & \textbf{68.0}  \\
    \bottomrule
    \end{tabular}
    \caption{Quantitative results on VSI-bench. vsGRPO-T, vsGRPO-O, and vsGRPO-V refer to GRPO training on VSI-100k with prompts of \text{\textcolor{red}{think-mode}}, \text{\textcolor{green}{observe-mode}}, and \text{\textcolor{blue}{vanilla-mode}} respectively. \textcolor{blue}{V}, \textcolor{red}{T}, and \textcolor{green}{O} in the Eval. 
    Mode column refer to using \textcolor{blue}{vanilla-mode}, \textcolor{red}{think-mode}, and \textcolor{green}{observe-mode} prompts for evaluation, respectively.
    We also present the best performance of open-source models under the specific model size, like LLaVA-NeXT-Video~\citep{li2024llava} (LNV for short) and InternVL2~\citep{chen2024internvl} (IVL2 for short), and close-source ones like GPT-4o~\citep{hurst2024gpt} and Gemini-1.5 Pro~\citep{team2024gemini}.}
    \label{tab:baselines}
\end{table}

\textbf{Experimental Settings.} 
Unless specified otherwise, we use Qwen2-VL-2B/7B as the base models due to resource constraints. 
We employ LoRA~\citep{hu2022lora} training with a learning rate of $10^{-5}$ for Qwen2-VL-2B and $5\times10^{-6}$ for Qwen2-VL-7B.
We conduct 14 rollouts per question and set the default sampling temperature to 1. 
The KL divergence coefficient $\beta$ is set to 0.0001.

\subsection{Results and Analyses}

\subsubsection{Main Results}
Let vsGRPO-T, vsGRPO-O, and vsGRPO-V denote the GRPO training on VSI-100k with prompts of \text{\textcolor{red}{think-mode}}, \text{\textcolor{green}{observe-mode}}, and \text{\textcolor{blue}{vanilla-mode}} respectively. 
We evaluate them with the corresponding test prompts by default. 
Given the studies in the previous section, we also test the trained models with {\color{blue}vanilla-mode} prompts.

As shown in Table~\ref{tab:baselines}, for models based on Qwen-VL-2B, all GRPO fine-tuned models improve over the baseline.
Besides, for the models trained with CoT prompting strategies, their CoT test performance outperforms vanilla one. 
This indicates that GRPO training can effectively enhance the model's long reasoning capabilities.
Notably, directly applying the {\color{blue} vanilla-mode} prompting strategy yields the best performance improvements, particularly for NA questions, and even outperforms GPT-4o.
We refer to this model as \textbf{vsGRPO-2B} by default. 
This underscores the conclusion that CoT prompting is ineffective for the small-sized Qwen2-VL-2B on the VSI-bench.

In terms of Qwen2-VL-7B, we only tried vsGRPO-V considering the above results. 
We observe that vsGRPO-V performs the best on two subtasks—object counting and absolute distance. 
Moreover, the test performance on the Route Plan is also improved, similar to the 2B case.
This is possibly because the Route Plan can be divided into sub-tasks that include relative direction, indicating inter-task generalization. 
With only 7B model size, we note that our model shows performance comparable to that of the leading open-source model, LLaVA-NeXT-Video-72B~\citep{li2024llava}. 

\subsubsection{Importance of KL Penalty}
\begin{figure}[t]
    \centering
    \includegraphics[width=0.45\linewidth]{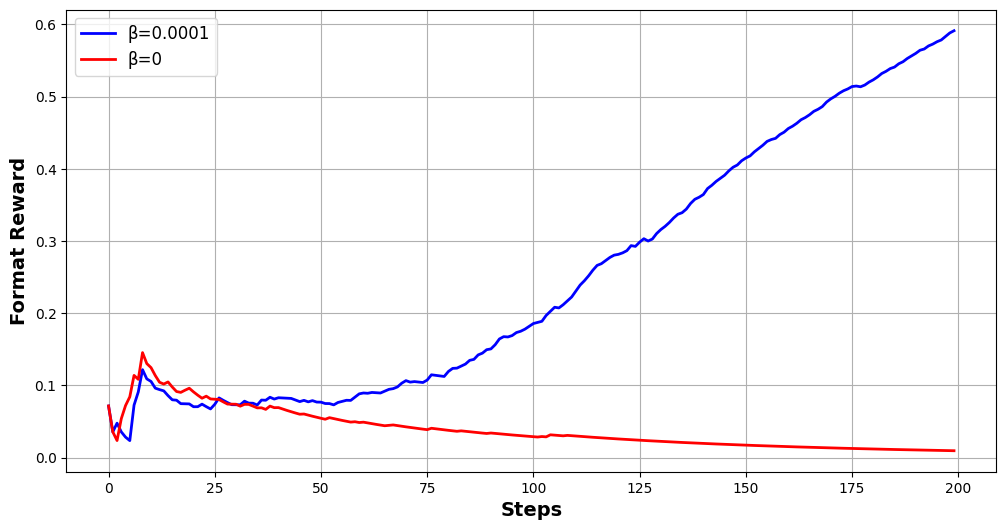}
    \includegraphics[width=0.45\linewidth]{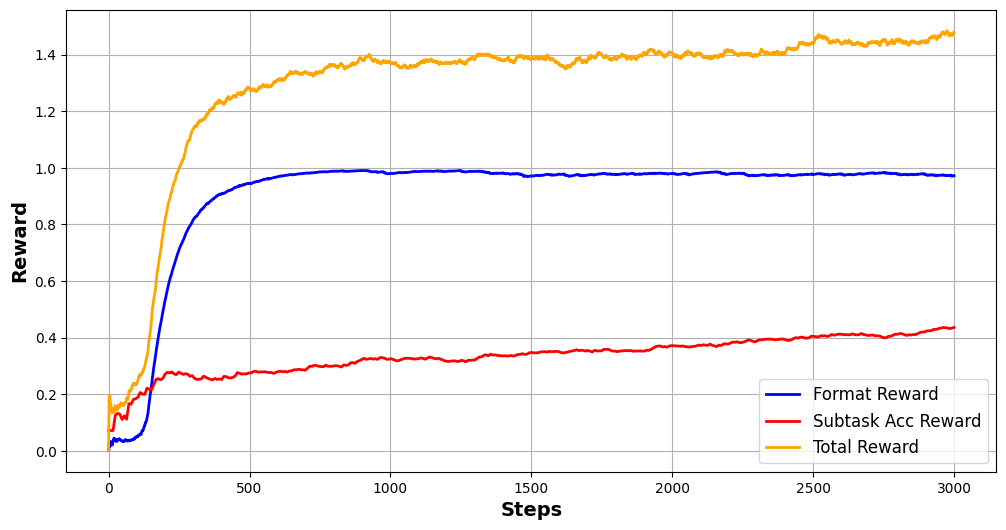}
    
    \caption{Left: the format reward curve of $\beta=0$ and $\beta=0.0001$ during training. Right: the curve of the format reward, the accuracy reward curve of one subtask of VSI-100k, and the total reward curve during GRPO training.}
    \label{fig:dicussion}
\end{figure}

The KL penalty term plays a crucial role in regulating the divergence between the online policy and the frozen reference one.
It avoids the model straying too far from the initial point. 
While some works~\citep{yu2025dapo,meng2025mm} advocate for removing the KL penalty to enhance performance, we have observed that doing so can easily lead to training collapse, as illustrated in Figure~\ref{fig:dicussion} (left). 
In contrast, introducing a positive $\beta$ (even very small, such as $0.0001$) can effectively address this issue. 
This may be attributed to the specific nature of VSI reasoning problems.

\subsubsection{Reward Hacking}
During the training, we observed that the model sometimes finds ways to obtain high rewards that do not align with our intentions.
One example is that when training with observe-mode, there are some extreme samples in the rollouts, e.g., \texttt{<think> </think> <answer>xx</answer>}.
While this format is technically correct, it degrades to a missing observation, a similar phenomenon also noted in VisualThinker-R1-Zero~\citep{zhou2025r1}.
Observing this, we opt to incorporate a length reward function for mitigation.
However, we note that some new generations just add extra \texttt{<think></think>} and \texttt{<answer></answer>} tags to exploit the length reward, which does not contribute to a meaningful thinking process. 
So, more reasonable reward functions should be explored.

\subsubsection{Dynamics of Various Rewards}
As shown in Figure~\ref{fig:dicussion} (right), during the GRPO training, the format reward converges to 1 quickly in the early stage, while the accuracy reward increases slowly.
It seems that there is an upper bound on the accuracy reward.
How to address this remains an open problem. 

\subsubsection{Comparison to Other Training Approaches}
We also compare our approach with commonly used fine-tuning algorithms, supervised fine-tuning (SFT) and direct preference optimization (DPO)~\citep{rafailov2023direct}, in Table~\ref{tab:baselines}. 
For SFT, we directly use the constructed VSI-100k for tuning. 
For DPO, 
the correct answer is modified to a wrong one to serve as the less-preferred answer. 

As shown, the two approaches both improve over the base model on the VSI-bench, but still lag behind GRPO-V.
Besides, the improvement of DPO is minor, which is perhaps because of the sub-optimal preference pair construction.

\section{Conclusion}
In this work, we center on the video-based visual-spatial intelligence of MLLMs. 
Using Qwen2-VL as the base models, we identify that visual-spatial reasoning capacities of Qwen2-VL-2B/7B cannot be activated via CoT prompts.
We construct \textbf{VSI-100k} to combat data scarcity and adapt GRPO training.
Extensive experiments demonstrate that \texttt{vsGRPO-2B} and \texttt{vsGRPO-7B} outperform models of the same size, highlighting the superiority of the GRPO approach in comparison to SFT and DPO.
We also share some findings for future research.

\bibliography{colm2025_conference}
\bibliographystyle{colm2025_conference}


\end{document}